# Problem Formulation as the Reduction of a Decision Model*

**David E. Heckerman**   **Eric J. Horvitz**
Medical Computer Science Group
Room 215, MSOB, Stanford Medical Center
Stanford, California 94305-5479

## Abstract

In this paper, we extend the QMR-DT probabilistic model for the domain of internal medicine to include decisions about treatments. In addition, we describe how we can use the comprehensive decision model to construct a simpler decision model for a specific patient. In so doing, we transform the task of problem formulation to that of narrowing of a larger problem.

## 1 Introduction

The structuring of problems that serve as the basis for inferential analysis has been considered one of the most ill-characterized phase of machine reasoning [17] Several investigators have speculated that we may never identify principled machinery for problem formulation [18, 3]. In this paper, we describe a formal approach to problem formulation in which we reduce a large decision model. In particular, we develop a comprehensive influence diagram for the domain of internal medicine. The model provides recommendations for treating a patient, given observations about that patient. To simplify the solution of the comprehensive model, we prune nodes and arcs from the model, based on observations that are specific to a given patient. We conjecture that the patient-specific influence diagram produces treatment recommendations that are typically identical to those recommendations derived from the comprehensive influence diagram.

*This work was supported by the National Library of Medicine under Grant RO1LM04529, by the National Science Foundation under Grant IRI-8703710, by a NASA Fellowship under Grant NCC-220-51, and by the U.S. Army Research Office under Grant P-25514-EL.

## 2 Related Research on Problem Formulation

Previous artificial-intelligence approaches to problem formulation—both heuristic and formal—involve the direct synthesis of problem-specific models rather than the reduction of comprehensive models. Researchers have made the assumption (typically implicit) that computer-based reasoning systems must be limited to the construction of relatively simple models for action at run time. A heuristic approach to problem formulation can be found in the Present Illness Program (PIP), developed over a decade ago [13]. PIP was designed to assist physicians with patients presenting with different types of swelling. All possible hypotheses considered by PIP, are stored in the system's *long-term memory*. Each disease in the knowledge base is associated with a set of observable criteria, called *triggers*, that are used to make decisions about whether a disease should be considered as *active*. Information about competing and complementary diseases is also stored in the long-term memory. A disease is *activated*, or brought into consideration in PIP's *short-term memory*, when a trigger is observed. Diseases that are competitors or complements also are brought into consideration as *semi-active* hypotheses. A problem, composed of active and semi-active hypotheses, is addressed with logical and quasi-probabilistic analyses in the working memory.

Researchers have pursued heuristic problem-formulation methodologies primarily because they reduce the computational burden by selecting a subset of distinctions for analysis. These heuristic approaches, however, are intrinsically limited by their poor characterization. In recent work, spanning artificial intelligence and decision science, several investigators have developed formal approaches that automate the formulation of decision problems. For example, Wellman has examined the identification of tradeoffs through utility-dominance theorem proving [21]. Also, Holtzman [8] and Breese [1] have developed rule-based techniques for constructing influence diagrams. In Section 7, we compare our approach for problem formulation to these approaches in the context of the internal-medicine domain.

## 3 Beyond Internist-1: QMR and QMR-DT

Our work is motivated by the QMR-DT, a decision-theoretic decision-support system for internal medicine, based on the comprehensive Quick Medical Reference (QMR) knowledge base [7, 5, 16]. The QMR reasoning system is the primary descendant of the Internist-1 project at the University of Pittsburgh [15, 12]. The QMR knowledge base was developed, and is being refined, by Miller and other researchers [11]. Over 25 person-years of effort have been directed at the construction of the QMR knowledge base.

QMR (and Internist-1) relies on heuristic numeric weighting schemes for reasoning under uncertainty. Several years ago, one of us developed a mapping between these methods and probability theory [6]. That early work, and more recent work within the QMR-DT (for QMR decision-theoretic) research group,[1] led to a reformulation of the QMR knowledge base in the form of a belief network for internal medicine.

Figure 1 portrays the general structure of the current QMR-DT belief network. Each disease in the upper layer of the network conditions a subset of manifestations in the lower layer of the network. The actual belief network contains 534 diseases, 4040 manifestations, and a 40740 arcs. Both diseases and manifestations are binary or two-valued distinctions. Also, diseases are marginally independent, and features are conditionally independent given an instance of diseases (i.e., given an assignment of *true* or *false* to each disease). Although not shown in the network, we model the influence of multiple diseases on a single manifestation using the assumption of *causal independence*. In particular, we use Pearl's *noisy OR-gate* model [14]. In the future, we plan to extend this model to include nonbinary diseases and manifestations, dependencies among diseases and manifestations, intermediate pathophysiological states, and manifestations that condition diseases.

The QMR-DT group is currently developing algorithms for the computation of the marginal poste-rior probability of each disease given instances of one or more of the manifestation nodes. Although this computation is an NP-hard task [4], preliminary results with several algorithms are encouraging. One algorithm, called Quickscore, performs inference efficiently when the number of findings observed to be present in a patient is small [5]. Another algorithm determines bounds on disease probabilities; these bounds shrink as the amount of computation increases [7]. An algorithm based on Monte-Carlo sampling, provides approximate disease probabilities; the approximations improve as sampling progresses [16].

---

[1]The QMR-DT team also includes Gregory Cooper, Max Henrion, Harold Lehmann, Blackford Middleton, Randolph Miller, and Michael Shwe.

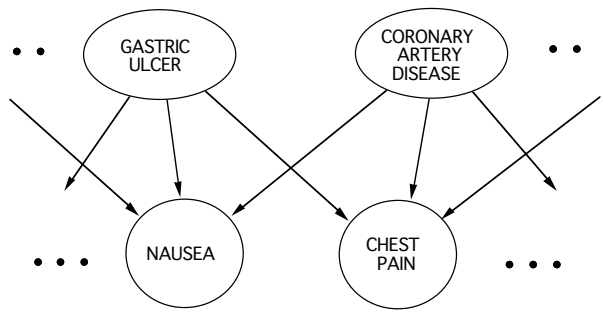

Figure 1: A portion of the QMR-DT belief network. The upper layer of the network consists of over 500 disease propositions. These propositions are associated with over 4,000 manifestations represented as nodes in the bottom layer of the network.

## 4 From Belief to Action

Determining the probability that single diseases or disease combinations have manifested in a patient is only one component of the internal-medicine problem. In addition, an automated reasoner should consider the possible actions that a physician might take, and the desirability of the possible consequences of those actions. For example, suppose an expert system determines that the probability that a given patient has syphilis is small—say, 0.01. If the program only considers the likelihoods of diseases, it will probably fail to suggest to a physician user that he should consider the disease. This failure to suggest the disease to the physician, however, is probably inappropriate, because the side effects of the treatment for syphilis (penicillin) are minimal, and the consequences of the untreated disease are severe.

In the domain of internal medicine, physician actions include treatments and other patient interven-tions, and the performance of tests to gather additional information about a patient. In this paper, we extend the QMR-DT model to include treatment actions only. This extension is illustrated by a portion of an influence diagram for internal medicine shown in Figure 2. In the diagram, CORNEAL HERPES ZOSTER and CORNEAL HERPES SIMPLEX are two diseases that can cause RED EYE. ACYCLOVIR and PREDNISONE are two treatments for CORNEAL HERPES ZOSTER, whereas ARA-A is a treatment for CORNEAL HERPES SIMPLEX. In general, there can be no treatment, one treatment, or more than one treatment for a given disease. Also, each treatment option may be associated with many (possibly an infinite) number of option alternatives, although in this paper, to keep the discussion simple, we limit the number alternatives for each treatment to two: *true* and *false*. The diamond node $u$ represents the overall utility of the patient. The diamond nodes $u_1$, $u_2$, and $u_3$, called *subvalue nodes* [19, 20], represent components of overall patient utility. The nodes $u_1$ and $u_3$ encode the patient utilities associated with the treatment of CORNEAL HERPES ZOSTER and CORNEAL HERPES SIMPLEX, respectively. The node $u_2$ represents the fact that prednisone therapy can be extremely detrimental to a patient if he has CORNEAL HERPES SIMPLEX. We refer to the entire influence diagram for the QMR domain as the *comprehensive decision model*.

We measure the utilities underlying the comprehensive model using standard gambles. For example, let us consider the utility associated with the node $u_3$ in Figure 2 that corresponds to the situation where a patient has CORNEAL HERPES SIMPLEX and is not receiving ARA-A, the only treatment for that disease. To assess this utility, we ask the patient (or an agent of that patient), "Imagine that there is a magic pill that will cure your CORNEAL HERPES SIMPLEX without any side effects with probability $p$, but will kill you immediately and painlessly with probability $1 - p$. What probability $p$ makes you indifferent between taking the pill and remaining in your current situation?" As another example, let us consider the utility that corresponds to the situation where the patient does not have CORNEAL HERPES SIMPLEX, and is erroneously receiving ARA-A. To assess this utility, we ask the patient "Imagine that there is a magic pill that will eliminate all side effects of ARA-A with probability $q$, but will kill you immediately and painlessly with probability $1 - q$. What probability $q$ makes you indifferent between taking the pill and remaining in your current situation?" Similarly, we can assess the remaining two utilities associated with $u_3$. Most people find these utilities difficult to assess directly, especially when the magic-pill probabilities are small (e.g., $q$ in the previous example). Howard, however, has developed a model that greatly simplifies such assessments [10].

In constructing the comprehensive decision model for the QMR domain, we assume that the magic-pill probabilities are independent of other diseases that the patient may have and of other treatments that the patient may be receiving. The assumption is a good one, because we can always introduce nodes, as we introduced $u_2$, to represent interactions among diseases and treatments. In making this assumption of independence, we can select the treatment alternatives that maximize the expected utility of $u$ by choosing those alternatives that maximize the expected utility of each of the $u_i$ separately.[2] We thereby reduce the number computations that we require to solve the comprehensive decision model.

## 5 Problem Formulation as Large-Model Reduction

The independence assumption we just discussed is a large step toward tractable solutions of the comprehensive decision model. Nonetheless, for most patient cases, we shall still require inordinate amounts of time to solve the model. We address this difficulty by using observations about a given patient to build a *patient-specific decision model*. This decision model is likely to be smaller than the comprehensive decision model, and is also likely to give us the same recommendations for action at a lower computational cost. We view the construction of this model as problem formulation.

Before we discuss the construction, let us consider the quantity called $p_{ij}^*$, the lowest probability of disease $d_j$ such that treatment $a_i$ is warranted by the comprehensive model, given that the patient has at most disease $d_j$ and receives no other treatments. We compute $p_{ij}^*$ by setting all diseases except $d_j$ and all treatments except $a_i$ to false, and then solving the comprehensive decision model. The computation does not depend on specific observations about a patient. Thus, we have to compute each $p_{ij}^*$ only once for the comprehensive model.

Given the $p_{ij}^*$, we can construct a patient-specific decision model in two steps. First, given disease manifestations $\phi$ for a particular patient, we compute the marginal posterior probability of each disease $d_j$, denoted $p(d_j|\phi)$. If the number of positive findings in $\phi$ is small, we can use the Quickscore algorithm to perform this inference. Otherwise, we can use Henrion's algorithm that provides bounds on these probabilities. In either case, we instantiate $a_i$ to false if and only if, for every disease $d_j$ that has $a_i$ as a possible treatment, the upper bound for the probability of $d_j$ falls below $p_{ij}^*$ for that treatment–disease pair. Second, we discard the portions of the comprehensive decision model that are not relevant to the utility of the patient, given those decision variables that we have instantiated. A simple algorithm for doing so is as follows: (1) for all subvalue nodes $u_i$, if all decision-node predecessors of $u_i$ are set to false, then remove $u_i$ from the comprehensive model; (2) remove all chance and decision nodes that become disconnected from the node $u$.

---
[2]Such decomposition is possible for any utility model that is additive or multiplicative [9]. The standard gamble model that we describe here is multiplicative.

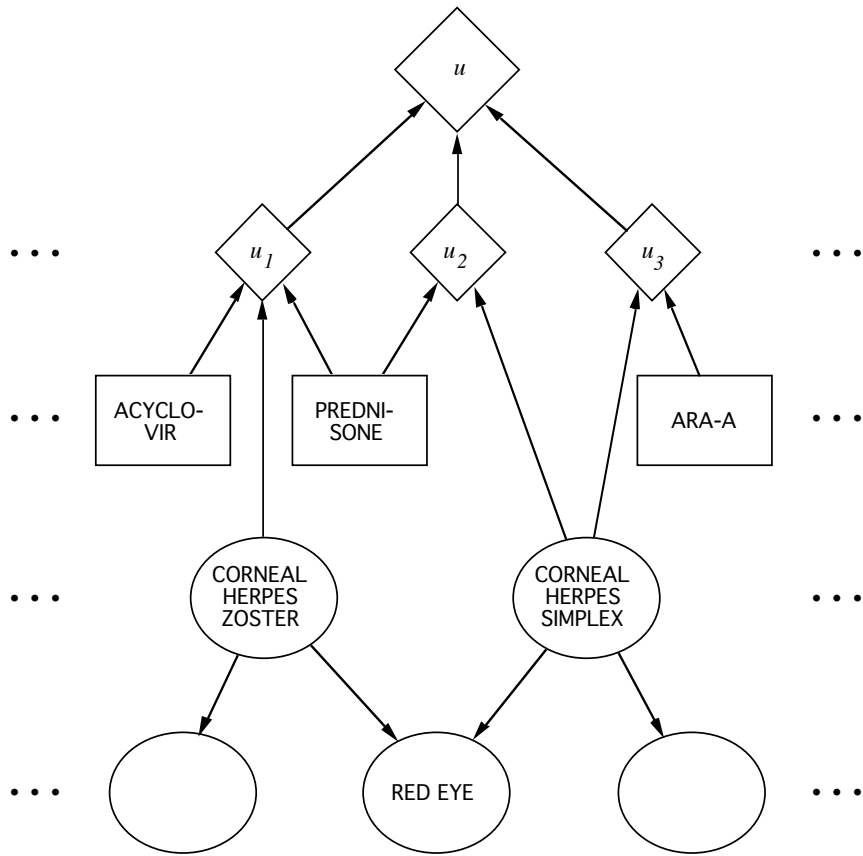

Figure 2: A portion of the comprehensive decision model for the QMR domain. CORNEAL HERPES ZOSTER and CORNEAL HERPES SIMPLEX are two diseases that can cause RED EYE. ACYCLOVIR and PREDNISONE are two treatments for CORNEAL HERPES ZOSTER, whereas ARA-A is a treatment for CORNEAL HERPES SIMPLEX. The overall utility of the patient is represented by the the diamond node labelled $u$. The diamond nodes $u_1$, $u_2$, and $u_3$, called *subvalue nodes*, represent components of patient utility that contribute independently to the overall utility of the patient. For example, the node $u_2$ represents the fact that prednisone therapy can be extremely detrimental to a patient if he has Corneal Herpes Simplex.

The solution of a patient-specific decision model is likely to be significantly more tractable than that of the comprehensive decision model. One obvious simplification is that a patient-specific decision model will probably contain fewer nodes than the comprehensive decision model. Another, more important, simplification is illustrated by the portion of comprehensive decision model shown in Figure 2. In particular, suppose our problem-formulation procedure instantiates PREDNISONE to false for a particular patient. In this case, we can solve the ACYCLOVIR and ARA-A portions of the influence diagram separately. A similar simplification is illustrated by the portion of the comprehensive decision model in Figure 3. If the problem-formulation procedure instantiates THEOPHYLLIN to false, then we can determine the decisions of whether or not to administer DIGOXIN and ERYTHROMYCIN independently.

In general, we say that a procedure for pruning a comprehensive decision model is *sound* if and only if the treatment decisions that are determined by the process of pruning the large model and by solving the reduced model are identical to those decisions determined by solving the comprehensive model. Our approach is not sound. For example, consider the case where $a_i$ is a treatment for diseases $d_j$ and $d_k$, but has many side effects. Let us suppose that, for a given patient, the upper bounds for $p(d_j|\phi)$ and $p(d_k|\phi)$ fall just below $p_{ij}^*$ and $p_{ik}^*$, respectively. In this situation, our approach instantiates $a_i$ to false. Within the comprehensive decision model, however, the treatment $a_i$ may be optimal, because the positive benefits of $a_i$ applied to both diseases may outweigh the negative side effects of the treatment. Nonetheless, we conjecture that such situations will arise rarely. After we construct a substantial subset of the comprehensive decision model for the QMR domain, we shall test this conjecture.

We have described our approach for creating a patient-specific decision model in terms of reducing a large comprehensive decision model by instantiating particular treatment nodes. Alternatively, as illustrated in Figure 4, we can view this approach as a *model-construction process*. From this perspective, treatment nodes that are not instantiated to false become *active*. In addition, subvalue and chance nodes that are relevant to $u$, given the instantiated treatment nodes, become *active*. The patient-specific decision model is then constructed from these active nodes and from the relationships among these nodes found in the comprehensive decision model.

## 6 Problem Formulation as a Metalevel Decision Analysis

We can view our approach as a metalevel decision analysis in which we tradeoff the comprehensiveness of the model with the amount of time required to solve the model. In particular, we have assumed that (1) the cost of including a treatment in a model, given that a full analysis would have proven the treatment should not be undertaken, is small, and (2) the utility of omitting a treatment from a model, given that a full analysis would have indicated the treatment be performed, is large. The first assumption requires the cost of computation to be small, or that the addition of another distinction (or set of distinctions) does not increase the complexity of the analysis significantly. Additional work is needed to measure such metalevel costs and benefits more accurately, and to develop procedures for constructing patient-specific models that exploit these measurements once they are made.

## 7 Comparison to Other Formal Approaches

We believe that the rule-based and qualitative-tradeoff approaches to problem formulation are inadequate for handling the construction of patient-specific decision models in the QMR domain. In applying the rule-based approach, an expert would not build a comprehensive decision model. Instead, he would generate rules that correspond to small components of the influence diagram for the QMR domain. For example, one rule might describe the THEOPHYLLINE–DIGOXIN interaction in Figure 3. The primary drawback of this approach—when applied to internal medicine—is that it will be extremely difficult for an expert to construct a large set of rules that are *consistent*. The lack of consistency is a limitation, because there is no guarantee that influence diagrams constructed by the rule-based approach will be valid unless the rules used to build those diagrams are self consistent [2]. Our approach does not suffer from this drawback, because the expert builds an influence diagram for the entire QMR domain explicitly, and thereby guarantees the self consistency of the knowledge represented in that diagram. Of course, the rule-based approach may be useful in domains where it is impossible or extremely difficult to build an influence diagram for the entire domain. For example, the approach may be useful for stock-market trading, where unforeseeable situations arise frequently.

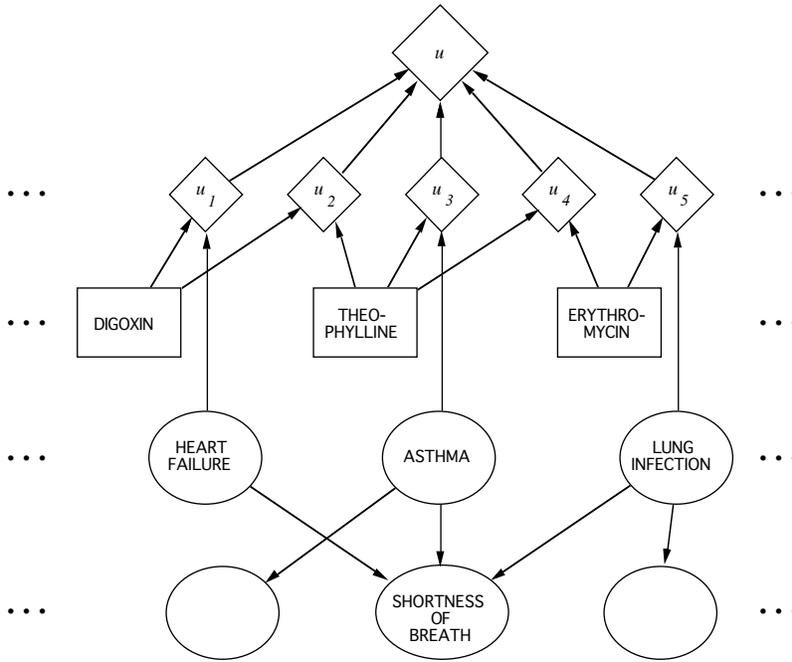

Figure 3: Another portion of the comprehensive decision model for the QMR domain. HEART FAILURE, ASTHMA, and LUNG INFECTION are diseases that can cause SHORTNESS OF BREATH. DIGOXIN is a treatment for HEART FAILURE, THEOPHYLLINE is a treatment for ASTHMA, and ERTHYROMYCIN is a treatment for LUNG INFECTION. THEOPHYLLINE interacts negatively with both DIGOXIN and ERYTHROMYCIN. The subvalue nodes $u_2$ and $u_4$ represents the negative interactions.

Wellman's approach examines only qualitative interactions among alternatives, beliefs, and preferences. For example, using the approach, we can represent the fact that THEOPHYLLINE might relieve the symptoms of ASTHMA, but we cannot represent the *degree* of belief that these symptoms will be relieved, nor can we represent the patient's *degree* of preference for such relief. Given these qualitative interactions, and a set of observations about a patient, Wellman's method can identify all decision, chance, and utility variables that *might be* relevant to the utility of that patient. Unfortunately, almost all variables in the QMR domain might be relevant to a patient's utility at the level of qualitative analysis. Thus, the approach approach will probably create extremely large patient-specific decision models. We need a more quantitative analysis to produce smaller patient-specific decision models whose solutions are tractable.

## 8  Summary

We have described an approach to problem formulation that uses the results of inference in a belief network to transform a large, comprehensive decision model into a smaller, patient-specific decision model. A crucial question is: Will this approach produce patient-specific decision models whose solutions are both accurate and tractable? We are currently implementing our approach within the QMR-DT framework to answer this question.

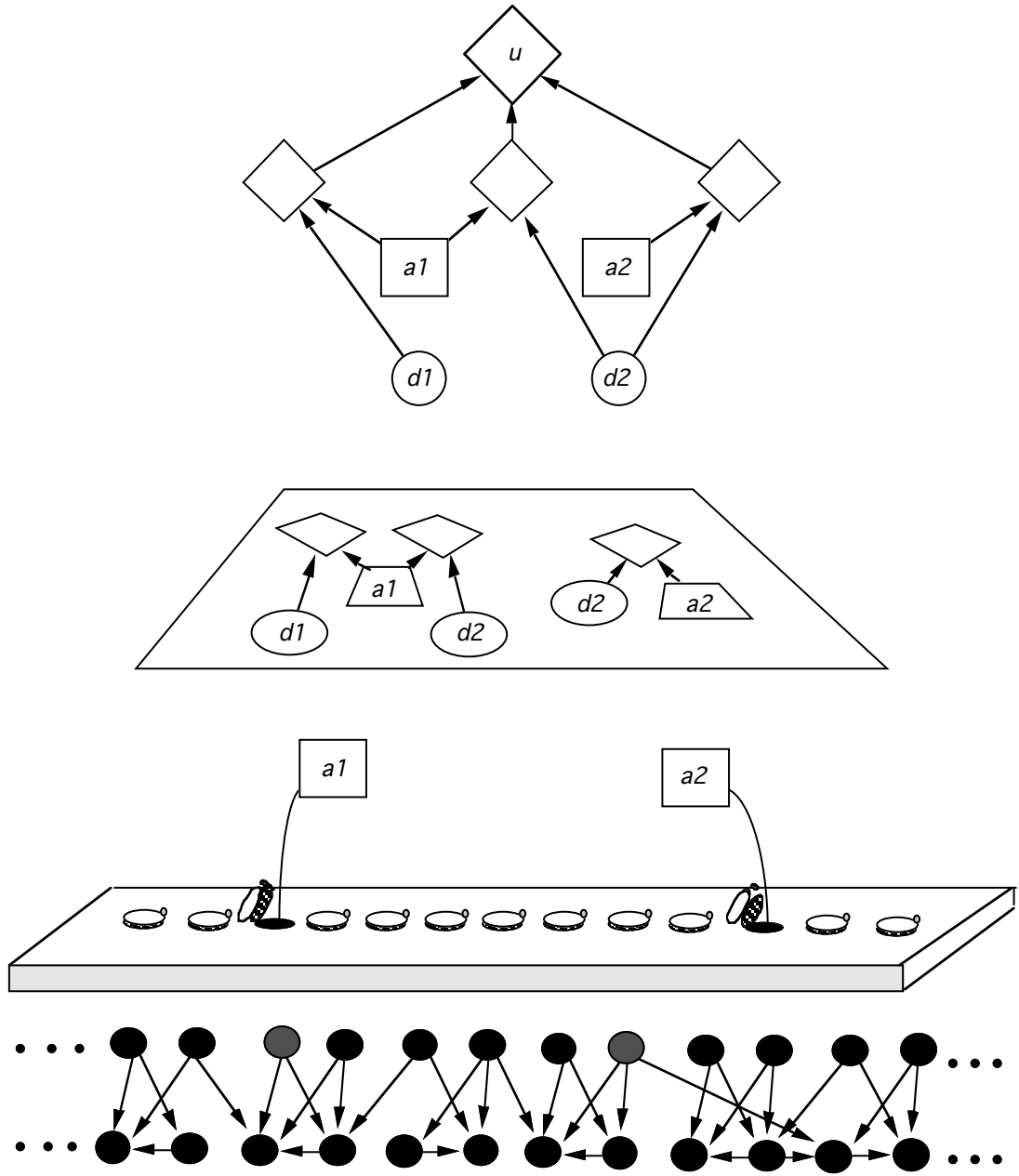

Figure 4: The construction of a patient-specific decision model. First, we apply an inference algorithm to the QMR-DT belief network that produces probabilities or bounds on the probabilities of each disease. Second, we identify treatments that we need to consider. In particular, we make treatment $a_i$ *active* if and only if there is some disease $d_j$ for which $a_i$ is a possible treatment, such that the upper bound for the probability of $d_j$ exceeds $p_{ij}^*$ for that treatment–disease pair. Third, we identify active subvalue and chance nodes as those nodes that are relevant to the decision problem, given the active treatment nodes. The active treatment, subvalue, and chance nodes, and the relationships among these nodes, form the patient-specific decision model, shown at the top of the figure. Nodes that represent disease findings are omitted from the figure.